\documentclass[runningheads]{llncs}

\usepackage{eccv}
\usepackage{eccvabbrv}

\usepackage{graphicx}
\usepackage{booktabs}
\usepackage{amsmath,amssymb}
\usepackage{url}
\usepackage{multirow}
\usepackage{array}
\usepackage{tabularx}
\usepackage{makecell}
\usepackage{float}
\usepackage{placeins}

\usepackage{hyperref}

\providecommand{\doi}[1]{\href{https://doi.org/#1}{doi:#1}}
\newcommand{\arxiv}[1]{\href{https://arxiv.org/abs/#1}{arXiv:#1}}

\newcommand{\Ind}{\mathbf{1}}
\newcommand{\eps}{\epsilon}
\newcommand{\DWConv}{\operatorname{DWConv}}

\newcommand{\sigmoid}{\operatorname{Sigmoid}}

\begin{document}
\mainmatter
\raggedbottom
\emergencystretch=2em

% ============================================================
\title{Controlled One-Click Benchmarking and Post-Fusion Recalibration for Support-Conditioned Medical Segmentation}
\titlerunning{One-Click Post-Fusion Recalibration}

\author{
Wang-Chi-Shiun\inst{1}
\and Wei-Huan Hu\inst{2}
\and Liu-En-Chi\inst{2}
\and Liu Zhi Shun\inst{2}
\and ChengYi Peng\inst{2}
\and Shin-Jye Lee\inst{1}
}

\authorrunning{Wang-Chi-Shiun et al.}

\institute{
National Yang Ming Chiao Tung University, Taiwan
\and
Chung Shan Medical University, Taiwan
}

\maketitle

% ============================================================

\begin{abstract}
Sparse-prompt medical segmentation is hard to compare when prompts, supports, thresholds, and splits differ. This paper uses a controlled one-click protocol: each test image receives one positive foreground click, training-only support, validation-only checkpoint selection, and no test-time threshold tuning. We propose \textbf{Support-Aware Recalibration (SAR)}, a late post-fusion recalibration block for MultiverSeg that combines large-kernel spatial gating with squeeze-and-excitation channel recalibration after support--query fusion. BUSI, NerveUS-BP-present, and Kvasir-SEG are evaluated with matched 697 foreground-present sample budgets, shared splits/clicks, and threshold-free Dice-AUC/IoU-AUC. Compared with a protocol-matched raw-support MultiverSeg baseline, the main SAR configuration shows favorable mean Dice-AUC/IoU-AUC performance on BUSI, NerveUS-BP-present, and Kvasir-SEG, while reducing logged training time in this implementation. These controlled results indicate that late post-fusion recalibration is useful under a leakage-controlled one-click protocol.
\end{abstract}

% ============================================================

\section{Introduction}

\begin{figure}[!tbp]
  \centering
  \includegraphics[width=0.70\linewidth]{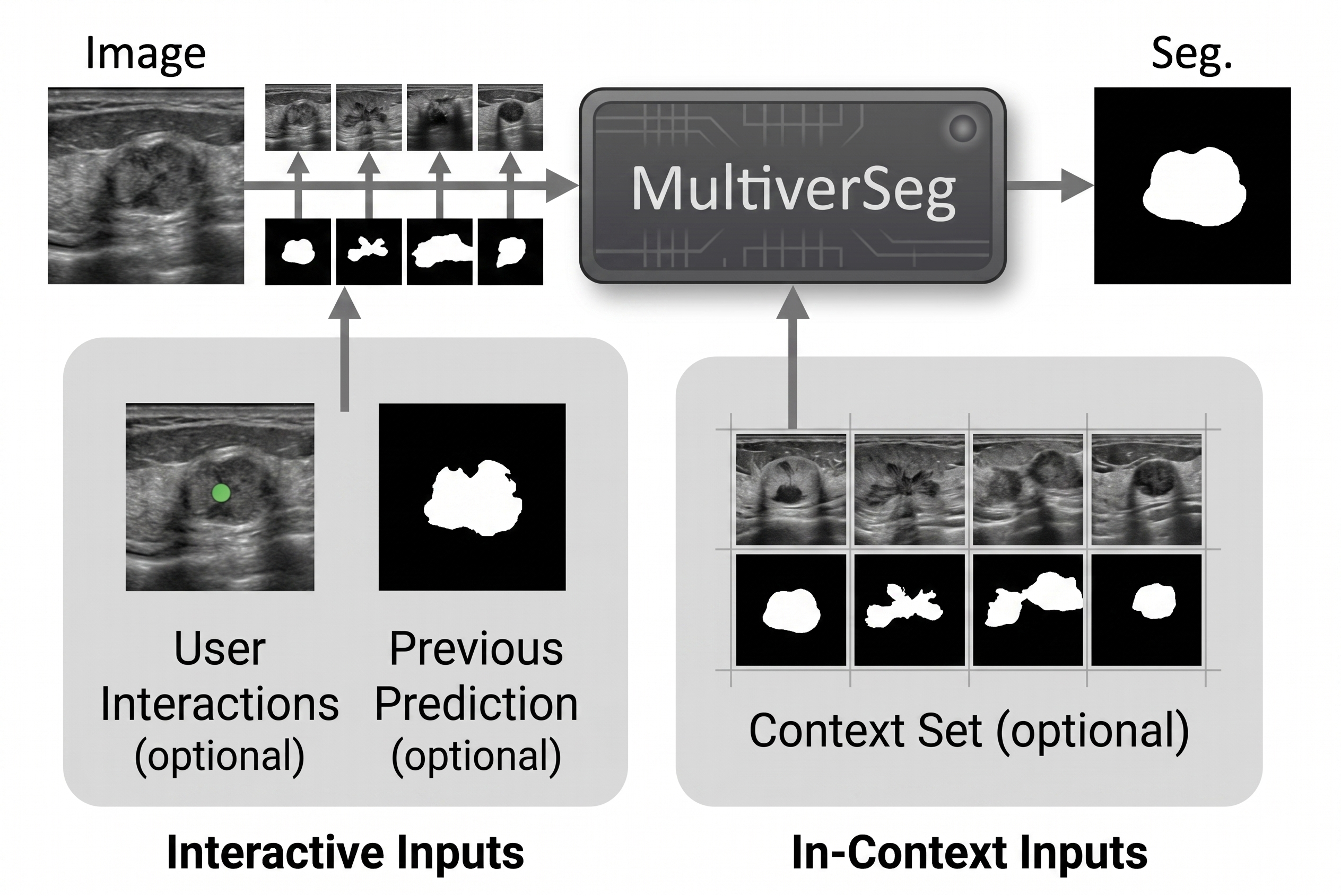}
  \caption{Overview of the baseline MultiverSeg paradigm~\cite{multiverseg}. The framework combines interactive inputs with an optional in-context support set to guide segmentation.Adapted from Wong et al.~\cite{multiverseg} under CC BY 4.0, with modifications.}
  \label{fig:multiverseg_base}
\end{figure}

Medical segmentation often relies on dense annotations, while promptable methods reduce annotation burden through clicks, boxes, scribbles, or iterative correction~\cite{grabcut,deepigeos,ritm,focalclick,simpleclick,sam,sam2}. We study a stricter setting: each test image receives exactly one positive foreground click, with no box, negative click, previous mask, threshold search, or test-time support selection. Figure~\ref{fig:multiverseg_base} illustrates the baseline MultiverSeg paradigm used in this controlled one-click protocol.

MultiverSeg is used because it combines query inputs with labeled support examples~\cite{multiverseg}. All support inputs are built only from the training split, so the architectural question is whether recalibration can improve the fused support-conditioned representation without changing the prompt budget or selecting support at test time.

Ultrasound segmentation is a useful stress case: one click gives location but not extent, and speckle, acoustic shadows, weak boundaries, and low contrast make local evidence unreliable~\cite{busi,samus}. We introduce Support-Aware Recalibration (SAR), a lightweight post-fusion block that applies a large-kernel gate and squeeze-and-excitation (SE) channel reweighting after the last two support--query fusion stages.

All MultiverSeg-based variants share the same splits, click lists, support source, fine-tuning policy, checkpoint-selection rule, and probability-map evaluation. The claim is limited to late post-fusion recalibration under this controlled protocol and to cross-dataset consistency, not to a universally optimal kernel size or insertion depth.

\medskip\noindent\textbf{Contributions.}
\begin{enumerate}
  \item \textbf{Controlled one-click protocol.} We study a one-positive-foreground-click setting with frozen splits, shared click lists, training-only support, validation-only checkpoint selection, and fixed evaluation rules without test-set threshold tuning.
  \item \textbf{Equal-data and support-source-controlled benchmark.} We control the sample budget and the training support source for BUSI, NerveUS-BP-present, and Kvasir-SEG~\cite{busi,nerveus,kvasirseg}, each capped to 697 foreground-present samples with identical 488/104/105 train/validation/test sizes. The benchmark separates compact support tensors (CSTs) from raw-support processing.
  \item \textbf{Late post-fusion Support-Aware Recalibration.} We evaluate SAR after support--query fusion, combining large-kernel spatial gating with SE channel reweighting. The main configuration and nearby diagnostics test kernel size and late insertion depth, rather than unrelated method families.
  \item \textbf{Threshold-free evaluation across three datasets.} We report three-seed threshold-free ablations with paired bootstrap confidence intervals on BUSI, NerveUS-BP-present, and Kvasir-SEG. The results separate raw-support processing, compact support construction, component-level SAR variants, and input/pre-fusion/post-fusion placement. Train time and parameter count are tabulated, and single-pass GFLOPs are reported for the main compact-support SAR versus raw-support comparison.
\end{enumerate}

\section{Related Work}

\medskip\noindent\textbf{Promptable and interactive segmentation.}
Classical interactive segmentation uses human input to constrain foreground extraction, with GrabCut as an early example~\cite{grabcut}. Deep methods learn from clicks, scribbles, or previous masks~\cite{deepigeos,ritm,focalclick,simpleclick}. SAM and SAM 2 scale promptable segmentation to broad image/video settings~\cite{sam,sam2}; medical variants such as MedSAM, SAMed, Medical SAM Adapter, SAMUS, SAM-Med2D, MA-SAM, Polyp SAM 2, and medical SAM 2 benchmarks show that clinical adaptation remains necessary~\cite{medsam,samed,medsa,samus,sammed2d,masam,polypsam2,sam2medical}. ScribblePrompt and nnInteractive study richer biomedical interaction~\cite{scribbleprompt,nninteractive}. U-Net and nnU-Net remain important supervised biomedical segmentation references~\cite{unet,nnunet}. We instead restrict the prompt budget to one positive click.

\medskip\noindent\textbf{Support-conditioned segmentation.}
Few-shot segmentation predicts a query mask from labeled support images. OSLSM, Prototypical Networks, PANet, PFENet, and DCAMA develop support/prototype aggregation strategies~\cite{oslsm,protonet,panet_fss,pfenet,dcama}; CD-ViTO shows that cross-domain few-shot performance is sensitive to support/instance quality and uses instance reweighting to form prototypes from weighted and averaged instance features~\cite{cdvito}. Following this reweighting idea, our CST uses foreground-mask weighting to aggregate a fixed training support bank. CD-ViTO uses learned detector-level instance scores, whereas our weighting is defined by segmentation masks and is used only to construct a compact support tensor. Building on UniverSeg's in-context medical segmentation line~\cite{universeg}, MultiverSeg extends support-conditioned guidance to scalable biomedical interactive segmentation~\cite{multiverseg}. We keep this backbone but force all support to be training-only and query-independent at validation/test time.

\medskip\noindent\textbf{Context modeling and recalibration.}
Transformers model long-range context~\cite{vit,swin,transunet,unetr,swinunetr}; large-kernel CNNs provide broad spatial context at convolutional cost~\cite{replknet,van,convnext}; state-space vision models offer another global modeling route~\cite{vmamba,mambavision}. SE, CBAM, and ECA recalibrate channel/spatial responses~\cite{senet,cbam,eca}. SAR applies these ideas after MultiverSeg support--query fusion, where support-conditioned foreground cues have already been formed.

\section{Method}

\subsection{One-Click Setting and Pipeline Overview}

Under the controlled one-click protocol, a query image $X\in\mathbb{R}^{H\times W\times C}$ and mask $Y\in\{0,1\}^{H\times W}$ provide one foreground click $(u,v)$. Validation/testing use the deterministic distance-transform center; training samples jittered high-distance foreground candidates. The click map is
\begin{equation}
M^+(i,j)=\Ind\left(|i-u|\le r_p,\ |j-v|\le r_p\right),
\qquad r_p=5.
\label{eq:prompt_map}
\end{equation}
All other prompt channels are null. The main controlled ablations report threshold-free Dice-AUC and IoU-AUC rather than a test-tuned threshold.

\begin{figure}[!tbp]
  \centering
  \includegraphics[width=0.78\linewidth]{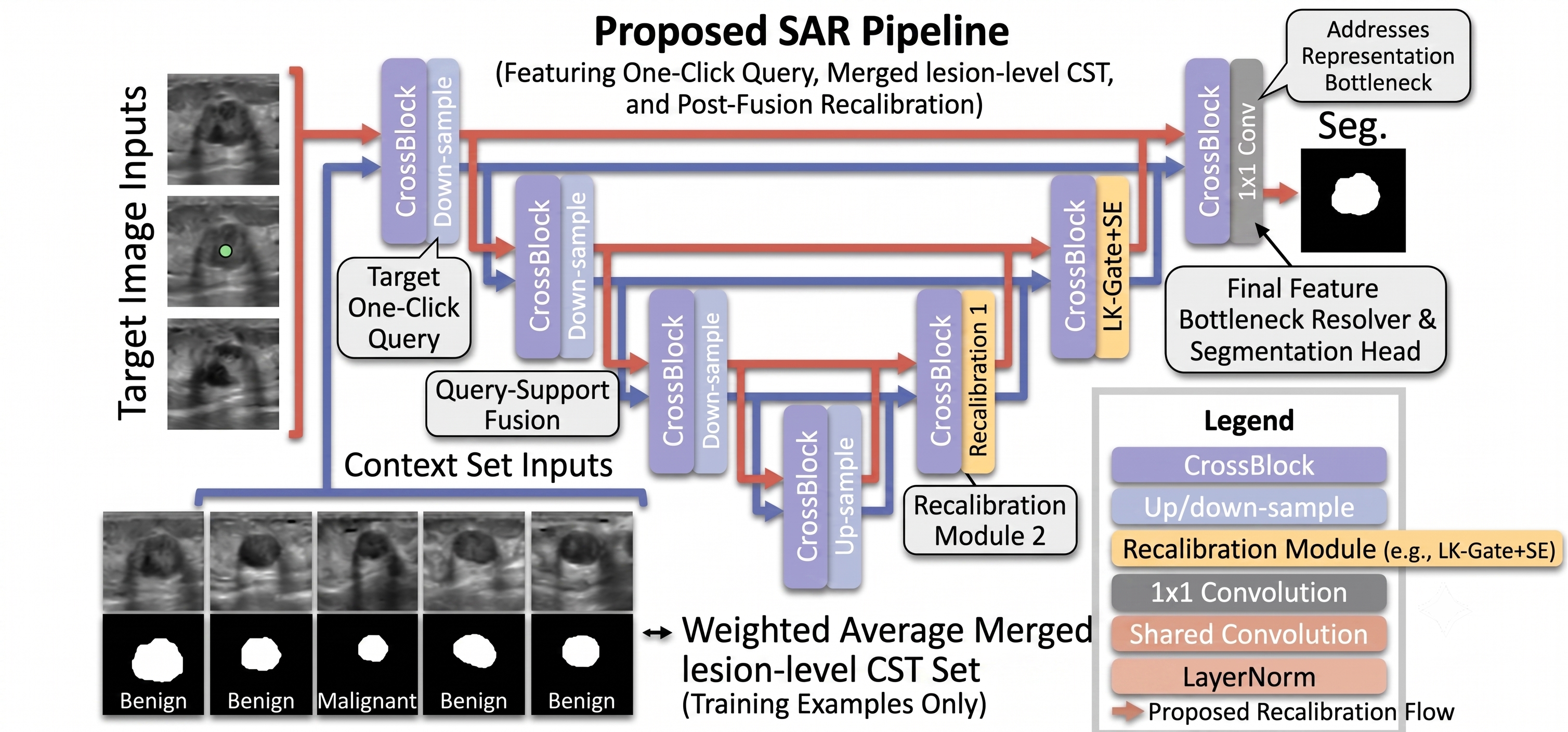}
  \caption{SAR pipeline. The target image and stamped foreground click form the query stream. The support stream uses a training-only compact support tensor (CST) built from a fixed 100-image training support bank: support images are averaged with foreground-mask weighting, and support masks are averaged over the same bank. The BUSI examples in the schematic are training-bank examples; in the implemented protocol, benign and malignant cases are merged into one lesion-level support pool before the CST is computed, and no validation/test class label is used for support selection. MultiverSeg CrossBlocks fuse query and support features. At the last two CrossBlock outputs, LK-Gate strengthens regional spatial coherence and SE recalibrates channels before the segmentation head.
  Adapted from Wong et al.~\cite{multiverseg} under CC BY 4.0, with modifications.}
  \label{fig:pipeline}
\end{figure}

Figure~\ref{fig:pipeline} shows SAR at inference. The CST is fixed before evaluation; MultiverSeg CrossBlocks fuse query and support streams; SAR then recalibrates the fused outputs from the last two CrossBlocks. The matched no-recalibration comparison isolates SAR under the same CST, prompt, and decoder.

Images are resized to $256\times256$ and normalized to $[0,1]$. BUSI and NerveUS-BP-present use ultrasound preprocessing; Kvasir-SEG uses the natural-image path. SAM-family baselines in Table~\ref{tab:main} follow their source-paper preprocessing and are included only as sparse-prompt references.

\subsection{Training-Only Support Construction}\label{sec:pseudo_support}

Let $\mathcal{T}$ be the training support pool; validation/test images and labels are never used to build support. After the dataset-specific pool is defined, each CST-based row uses a query-independent support bank $\mathcal{B}\subset\mathcal{T}$ with $|\mathcal{B}|=100$, sampled only from the capped training split for that dataset and seed. The bank is fixed before validation/testing and reused for every validation/test query. We set $\eps=10^{-7}$ in denominators and logarithms.

\medskip\noindent\textbf{BUSI support without test-class selection.}
BUSI labels are used only for split stratification and training-bank construction. Before validation/testing, benign and malignant training cases are merged into one query-independent lesion support pool:
\begin{equation}
\mathcal{T}_{\mathrm{BUSI}}=
\mathcal{T}_{\mathrm{benign}}\cup
\mathcal{T}_{\mathrm{malignant}} .
\label{eq:busi_merged_support}
\end{equation}
For BUSI, the 100-image support bank $\mathcal{B}$ is sampled after this merge from $\mathcal{T}_{\mathrm{BUSI}}$. The validation/test query label is never read to choose benign- or malignant-specific support.

The compact support tensor (CST) is computed from this fixed 100-image bank. Following CD-ViTO's support-instance reweighting~\cite{cdvito}, we use weighted support aggregation; here the weights are foreground masks, not learned detector instance scores:
\begin{equation}
\bar{S}=\frac{\sum_{(S_i,Y_i)\in\mathcal{B}}S_i\odot Y_i}{\sum_{(S_i,Y_i)\in\mathcal{B}}Y_i+\eps},
\qquad
\bar{Y}=\frac{1}{|\mathcal{B}|}\sum_{(S_i,Y_i)\in\mathcal{B}}Y_i.
\label{eq:mask_weighted_proto}
\end{equation}
Here $Y_i$ is broadcast over image channels and $\bar{Y}$ remains a soft support mask. This mask-weighted average acts as a segmentation-specific support prototype: foreground pixels from the 100-image bank contribute to the compact support image, while background texture is suppressed. Within each matched dataset/seed run, the CST is fixed before validation/testing, so the main SAR configuration and the no-recalibration CST baseline share the same support representation. Compact rows have $K=1$ at evaluation time, but the CST is not a single selected training image; it is a foreground-mask-weighted aggregate of 100 training-bank images. The 100-image bank is used only offline to estimate this fixed tensor and is not provided as 100 validation/test supports. Therefore, our ``support-source-controlled'' claim refers to the shared training support source, the capped training pool, and the absence of test-time support selection, not to an equal raw-shot count.

\medskip\noindent\textbf{Support representations.}
Compact-support MultiverSeg variants use one query-independent CST from Eq.~\eqref{eq:mask_weighted_proto}, built from the fixed 100-image training support bank. The raw-support comparator follows the original MultiverSeg usage more closely: it starts from the official MultiverSeg pretrained weights, is full-layer fine-tuned, and uses three query-independent raw training supports reused for every validation/test query. These three supports are sampled once from the capped training split, are not mask-weighted, and are not averaged into a compact tensor. This comparison is not claimed to be an equal-raw-shot support comparison; the matched no-recalibration comparison isolates SAR under the same CST support representation.

\subsection{CrossBlock Fusion and SAR}

CrossBlock is inherited from MultiverSeg~\cite{multiverseg}. It combines query features $Q^{(l)}$ with support/context features $C_k^{(l)}$ and aggregates support-conditioned responses:
\begin{equation}
G_k^{(l)}
=
\phi_{\mathrm{cross}}^{(l)}\left(Q^{(l)}, C_k^{(l)}\right),
\qquad
A^{(l)}=\frac{1}{K}\sum_{k=1}^{K}G_k^{(l)}.
\label{eq:crossblock_context_agg}
\end{equation}
The query representation is then updated by concatenating the query feature and aggregated support response:
\begin{equation}
F^{(l)}
=
\mathrm{LN}\left(
\rho\left(
\mathrm{Conv}^{(l)}\left([Q^{(l)},A^{(l)}]\right)
\right)
\right).
\label{eq:crossblock_update}
\end{equation}
where $\rho$ denotes the nonlinear activation function in the inherited MultiverSeg CrossBlock implementation.

SAR is applied after the last two CrossBlocks, where query appearance and training-only support cues have already been fused. Let $L$ denote the total number of CrossBlock stages. For each late fused map $F^{(m)}$, $m\in\{L-1,L\}$, SAR uses a large-kernel spatial gate with depthwise convolution (DWConv) and sigmoid activation $\sigma$:
\begin{equation}
\begin{aligned}
F_{\mathrm{sp}}^{(m)}
&=F^{(m)} \odot \sigma\left(Z^{(m)}\right),\\
Z^{(m)}
&=\mathrm{Conv}_{1\times1}^{(m)}
\left(\DWConv_{k\times k}^{(m)}\left(F^{(m)}\right)\right).
\end{aligned}
\label{eq:spatial_gate}
\end{equation}
followed by SE channel recalibration~\cite{senet}:
\begin{equation}
\tilde{F}^{(m)}=\mathrm{SE}^{(m)}\left(F_{\mathrm{sp}}^{(m)}\right),\qquad m\in\{L-1,L\}.
\label{eq:late_recalibration}
\end{equation}
\textbf{Support-Aware Recalibration (SAR).} The main SAR configuration uses LK-Gate+SE with $k=13$ on the last two CrossBlock outputs. Nearby diagnostics vary the kernel size, insertion depth, or remove one component to test whether the gain comes from late spatial gating, channel recalibration, or their combination.

\subsection{Training Objective and Implementation}

All variants based on MultiverSeg are fine-tuned end-to-end with the same composite segmentation loss. Let $z_i$ denote the logit at pixel $i$, $p_i=\sigmoid(z_i)$ the foreground probability, and $Y_i\in\{0,1\}$ the target mask. To reduce foreground--background imbalance, the BCE term uses an image-level positive weight:
\begin{equation}
\alpha
=
\mathrm{clamp}\left(
\frac{\sum_i (1-Y_i)}{\sum_i Y_i+\eps},
\alpha_{\min},\alpha_{\max}
\right),
\qquad
\alpha_{\min}=1,\quad \alpha_{\max}=10 .
\label{eq:pos_weight}
\end{equation}
The weighted BCE loss is
\begin{equation}
\mathcal{L}_{\mathrm{WBCE}}
=
-\frac{1}{HW}
\sum_i
\left[
\alpha Y_i\log(p_i+\eps)
+
(1-Y_i)\log(1-p_i+\eps)
\right].
\label{eq:wbce}
\end{equation}
The region-overlap terms are the soft Dice loss~\cite{vnet} and the soft IoU loss:
\begin{equation}
\mathcal{L}_{\mathrm{Dice}}
=
1-
\frac{2\sum_i p_iY_i}
{\sum_i p_i+\sum_iY_i+\eps},
\label{eq:soft_dice}
\end{equation}
\begin{equation}
\mathcal{L}_{\mathrm{SIoU}}
=
1-
\frac{\sum_i p_iY_i+\eps}
{\sum_i \left(p_i+Y_i-p_iY_i\right)+\eps}.
\label{eq:soft_iou}
\end{equation}
To encourage boundary agreement, let $P=\{p_i\}_{i=1}^{HW}$ denote the predicted probability map. A morphological edge map is computed with a $3\times3$ max-pooling operator:
\begin{equation}
\begin{aligned}
\mathcal{B}_3(U)
&=
\mathrm{clamp}\left(
\mathrm{MaxPool}_{3}(U)-\mathrm{MinPool}_{3}(U),
0,1
\right),\\
\mathrm{MinPool}_{3}(U)
&=-\mathrm{MaxPool}_{3}(-U).
\end{aligned}
\label{eq:morph_edge}
\end{equation}
With $p_i^b=\mathcal{B}_3(P)_i$ and $Y_i^b=\mathcal{B}_3(Y)_i$, the boundary Dice loss is
\begin{equation}
\mathcal{L}_{\mathrm{Bdry}}
=
1-
\frac{2\sum_i p_i^bY_i^b}
{\sum_i p_i^b+\sum_iY_i^b+\eps}.
\label{eq:boundary_dice}
\end{equation}
The final training objective is
\begin{equation}
\mathcal{L}
=
0.02\,\mathcal{L}_{\mathrm{WBCE}}
+
0.43\,\mathcal{L}_{\mathrm{Dice}}
+
0.40\,\mathcal{L}_{\mathrm{SIoU}}
+
0.15\,\mathcal{L}_{\mathrm{Bdry}} .
\label{eq:loss}
\end{equation}
Training uses AdamW~\cite{adamw}, learning rate $1\times10^{-4}$, weight decay $1\times10^{-4}$, cosine annealing, 65 epochs, batch size 1, gradient accumulation 2, and mixed precision. Checkpoints are selected by validation performance and evaluated on the test split. The three seeds $\{42,123,2024\}$ control initialization, shuffling, augmentation, training-time click jitter, and support-bank sampling for CST and raw-support rows; the split and validation/test clicks are fixed. Within a seed, matched variants share the same support bank whenever applicable. The intervals measure training-run variability under one split, not cross-split uncertainty.

Complexity uses parameter count and a single-pass forward GFLOP profile at $256\times256$. The main SAR configuration has $19.160$M parameters and $445.783$ GFLOPs, compared with $18.903$M and $409.293$ GFLOPs for the three-raw-support comparator. SAR is not a FLOP-level speedup. Its lower logged training time in this implementation reflects the full pipeline: the compact-support setting precomputes one CST once and reuses it, while the raw-support setting repeatedly loads and processes three supports with gradient-tracked computation, tensor movement, and backward-pass overhead.

\section{Experiments}

\subsection{Datasets, Splits, and Protocol}

We evaluate BUSI~\cite{busi}, NerveUS-BP-present~\cite{nerveus}, and Kvasir-SEG~\cite{kvasirseg}. BUSI normal images are excluded because a foreground click requires a lesion. The 697-sample cap is set by the BUSI lesion-present pool; NerveUS-BP-present and Kvasir-SEG are subsampled to match it, yielding identical 488/104/105 train/validation/test sizes.

\begin{table}[!htbp]
\centering
\small
\caption{Equal-data benchmark splits. BUSI uses its full lesion-present pool after excluding normal images; NerveUS-BP-present and Kvasir-SEG are subsampled to match the 697-sample BUSI foreground-present count.}
\label{tab:dataset_protocol}
\setlength{\tabcolsep}{4pt}
\begin{tabularx}{\linewidth}{>{\raggedright\arraybackslash}p{0.30\linewidth} c c c c c}
\toprule
\textbf{Dataset} & \textbf{Source pool} & \textbf{Used samples} & \textbf{Train} & \textbf{Val} & \textbf{Test} \\
\midrule
BUSI~\cite{busi} & lesion-present & 697 & 488 & 104 & 105 \\
NerveUS-BP-present~\cite{nerveus} & BP-present only & 697 & 488 & 104 & 105 \\
Kvasir-SEG~\cite{kvasirseg} & polyp-present & 697 & 488 & 104 & 105 \\
\bottomrule
\end{tabularx}
\end{table}

\begin{table}[!htbp]
\centering
\footnotesize
\caption{Control checklist. These controls reduce confounding from prompt strength, support construction, threshold handling, and test-set selection.}
\label{tab:leakage}
\renewcommand{\arraystretch}{1.18}
\setlength{\tabcolsep}{6pt}
\begin{tabularx}{\linewidth}{@{}>{\raggedright\arraybackslash\bfseries}p{0.22\linewidth} >{\raggedright\arraybackslash}X@{}}
\toprule
Factor & Control \\
\midrule
Prompt strength
& One positive foreground click only; no box, scribble, negative click, or previous mask. \\
\midrule
Support construction
& Support inputs for the models based on MultiverSeg are built only from training images before validation/testing. CST rows use a fixed 100-image bank only to precompute one support tensor; this is not used as 100 validation/test supports. The raw-support row uses three training supports. For BUSI, benign and malignant training cases are merged into one lesion support pool, and validation/test query labels are never used to choose support. \\
\midrule
Threshold handling
& No test-set operating point is selected. Thresholded scores use a fixed threshold, while controlled ablations use the same fixed threshold sweep for Dice-AUC/IoU-AUC. \\
\midrule
Prompt noise
& The same deterministic distance-transform validation/test click lists are reused across methods. \\
\midrule
Model selection
& Checkpoints are selected on the validation split only. \\
\midrule
Dataset size
& Each dataset uses 697 foreground-present samples and identical 488/104/105 splits; larger NerveUS/Kvasir pools do not receive extra training or validation data. \\
\midrule
Support pool
& Support is constructed or selected from the capped training split only; larger datasets cannot provide a larger or more diverse validation/test support pool. \\
\bottomrule
\end{tabularx}
\renewcommand{\arraystretch}{1.0}
\end{table}

Dice and IoU are computed from a binary mask $\hat{Y}_{\tau}$ obtained by thresholding the probability map at $\tau$:
\begin{equation}
\begin{aligned}
\mathrm{Dice}_{\tau}
&=\frac{2|\hat{Y}_{\tau}\cap Y|+\eps}{|\hat{Y}_{\tau}|+|Y|+\eps},\\
\mathrm{IoU}_{\tau}
&=\frac{|\hat{Y}_{\tau}\cap Y|+\eps}{|\hat{Y}_{\tau}\cup Y|+\eps}.
\end{aligned}
\label{eq:metrics}
\end{equation}
To reduce dependence on a single binarization threshold, the controlled ablations report threshold-free Dice-AUC and IoU-AUC:
\begin{equation}
\begin{aligned}
\mathrm{Dice\text{-}AUC}
&=
\frac{1}{\tau_{\max}-\tau_{\min}}
\int_{\tau_{\min}}^{\tau_{\max}}
\mathrm{Dice}_{\tau}\,d\tau,\\
\mathrm{IoU\text{-}AUC}
&=
\frac{1}{\tau_{\max}-\tau_{\min}}
\int_{\tau_{\min}}^{\tau_{\max}}
\mathrm{IoU}_{\tau}\,d\tau .
\end{aligned}
\label{eq:threshold_auc}
\end{equation}
In implementation, Eq.~\eqref{eq:threshold_auc} is approximated by averaging over the same fixed threshold grid $\mathcal{G}$ for every method and seed:
\begin{equation}
\widehat{\mathrm{Dice\text{-}AUC}}
=
\frac{1}{|\mathcal{G}|}\sum_{\tau\in\mathcal{G}}\mathrm{Dice}_{\tau},
\qquad
\widehat{\mathrm{IoU\text{-}AUC}}
=
\frac{1}{|\mathcal{G}|}\sum_{\tau\in\mathcal{G}}\mathrm{IoU}_{\tau}.
\label{eq:threshold_auc_discrete}
\end{equation}
When thresholded scores are needed, the threshold is fixed in advance or selected on validation only. Tables~\ref{tab:ablation}--\ref{tab:paired_ci} use threshold-free AUC with the same split, click list, support source, checkpoint rule, and threshold grid.

\subsection{Protocol-Level Medical SAM-Family Reference}
\begin{table}[H]
\centering
\scriptsize
\caption{Protocol-level sparse-prompt reference. SAM-family rows report source-paper fixed-threshold Dice/IoU; \textbf{Ours: SAR, controlled AUC} reports controlled threshold-free Dice-AUC/IoU-AUC and is not directly comparable. Only aligned Kvasir-SEG source rows are kept. Boldface in the SAR row highlights the controlled method row, not a direct superiority claim over source-paper baselines.}
\label{tab:main}
\setlength{\tabcolsep}{2.9pt}
\renewcommand{\arraystretch}{0.90}
\begin{tabularx}{\linewidth}{@{}>{\raggedright\arraybackslash}p{0.22\linewidth}
>{\raggedright\arraybackslash}X
r
c
c@{}}
\toprule
\textbf{Dataset} & \textbf{Method} & \makecell{\textbf{Reported}\\\textbf{Params. (M)}} &
\makecell{\textbf{Dice}\\\textbf{/ AUC}$\uparrow$} &
\makecell{\textbf{IoU}\\\textbf{/ AUC}$\uparrow$} \\
\midrule
\multirow{7}{*}{BUSI}
& SAM~\cite{sam} & 0.00 & 40.79 & 33.73 \\
& SAMed~\cite{samed} & 18.81 & 76.74 & 62.76 \\
& Med-SA~\cite{medsa} & 13.00 & 77.02 & 65.37 \\
& SAMUS~\cite{samus} & 130.10 & 77.81 & 66.18 \\
& MedSAM~\cite{medsam} & 93.73 & 75.96 & 61.80 \\
& HA-SAM~\cite{hasam} & 6.82 & 78.56 & 65.67 \\
& \makecell[l]{\textbf{Ours: SAR}\\\textbf{controlled AUC}} & \textbf{19.160} & \textbf{83.89} & \textbf{73.88} \\
\midrule
\multirow{7}{*}{NerveUS-BP-present}
& SAM~\cite{sam} & 0.00 & 46.76 & 38.53 \\
& SAMed~\cite{samed} & 18.81 & 78.80 & 67.32 \\
& Med-SA~\cite{medsa} & 13.00 & 78.12 & 66.39 \\
& SAMUS~\cite{samus} & 130.10 & 78.31 & 66.47 \\
& MedSAM~\cite{medsam} & 93.73 & 75.79 & 64.18 \\
& HA-SAM~\cite{hasam} & 6.82 & 79.44 & 66.73 \\
& \makecell[l]{\textbf{Ours: SAR}\\\textbf{controlled AUC}} & \textbf{19.160} & \textbf{81.09} & \textbf{69.89} \\
\midrule
\multirow{3}{*}{Kvasir-SEG}
& SAMUS~\cite{samus} & --- & 85.90 & 76.30 \\
& MedSAM~\cite{medsam} & --- & 86.20 & 79.50 \\
& \makecell[l]{\textbf{Ours: SAR}\\\textbf{controlled AUC}} & \textbf{19.160} & \textbf{88.52} & \textbf{81.26} \\
\bottomrule
\end{tabularx}
\renewcommand{\arraystretch}{1.0}
\end{table}

\begin{figure}[H]
  \centering
  \includegraphics[width=0.52\linewidth]{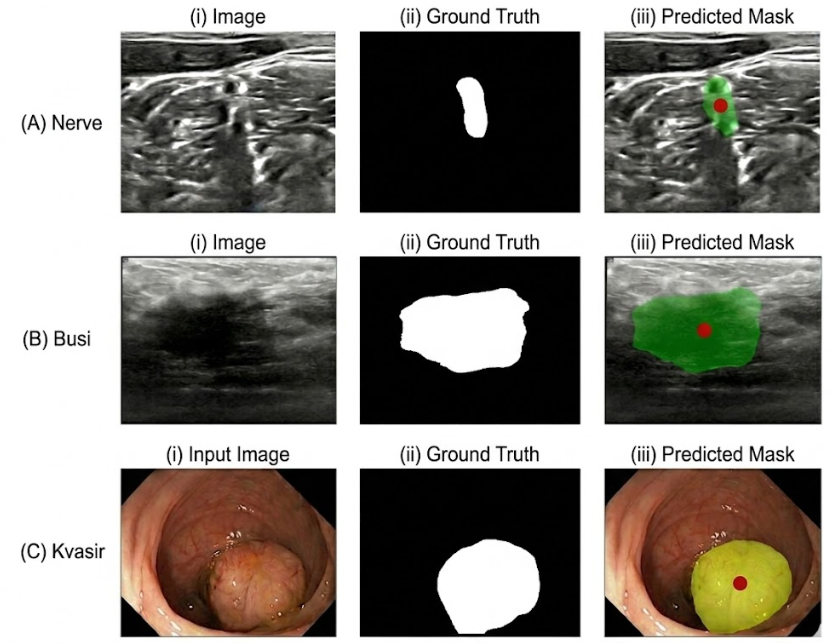}
  \caption{Qualitative one-positive-foreground-click examples on NerveUS-BP-present, BUSI, and Kvasir-SEG. Predictions use one positive foreground click, a training-only CST built from a fixed 100-image support bank, and a fixed binarization threshold $\tau=0.5$.}
  \label{fig:visual}
\end{figure}

Figure~\ref{fig:visual} shows qualitative one-click predictions under the same CST-based inference protocol.

\subsection{MultiverSeg-Controlled Threshold-Free Ablation Study}

\begin{table}[H]
\centering
\scriptsize
\caption{Controlled MultiverSeg-based ablation under threshold-free evaluation. Dice-AUC and IoU-AUC are reported in percent as mean $\pm$ standard deviation over three training seeds. \texttt{Sup.} denotes evaluation support; MW denotes one foreground-mask-weighted CST built from a fixed 100-image training support bank. Params are in millions, and Time is mean wall-clock training time per run in minutes in the matched implementation.}
\label{tab:ablation}
\setlength{\tabcolsep}{2.6pt}
\resizebox{\linewidth}{!}{%
\begin{tabular}{l c c c cc cc cc}
\toprule
\textbf{Variant} & \textbf{Sup.} & \textbf{Params} & \textbf{Time} &
\multicolumn{2}{c}{\textbf{BUSI}} &
\multicolumn{2}{c}{\textbf{Nerve}} &
\multicolumn{2}{c}{\textbf{Kvasir}} \\
\cmidrule(lr){5-6}\cmidrule(lr){7-8}\cmidrule(lr){9-10}
& & & & \makecell{\textbf{Dice}\\\textbf{-AUC}} & \makecell{\textbf{IoU}\\\textbf{-AUC}} & \makecell{\textbf{Dice}\\\textbf{-AUC}} & \makecell{\textbf{IoU}\\\textbf{-AUC}} & \makecell{\textbf{Dice}\\\textbf{-AUC}} & \makecell{\textbf{IoU}\\\textbf{-AUC}} \\
\midrule
A1: MW CST, no SAR & 1 MW & 18.903 & 59.8 & 83.38$\pm$0.14 & 73.24$\pm$0.30 & 80.24$\pm$0.31 & 68.92$\pm$0.46 & 87.79$\pm$0.64 & 79.95$\pm$0.93 \\
A2: MultiverSeg raw 3-shot & 3 raw & 18.903 & 89.5 & 83.00$\pm$0.45 & 72.77$\pm$0.50 & 80.12$\pm$0.57 & 68.96$\pm$0.76 & 88.02$\pm$0.74 & 80.32$\pm$0.95 \\
A3/SAR, $k=17$ & 1 MW & 19.221 & 81.9 & 83.85$\pm$0.23 & 73.79$\pm$0.31 & 81.06$\pm$0.46 & {69.94$\pm$0.69} & 88.63$\pm$0.12 & 81.38$\pm$0.02 \\
A4: SE only & 1 MW & 18.937 & 61.7 & 83.18$\pm$0.22 & 72.85$\pm$0.29 & 79.93$\pm$0.55 & 68.27$\pm$0.80 & 87.83$\pm$0.60 & 80.12$\pm$0.86 \\
A5: LK only, $k=13$ & 1 MW & 19.126 & 73.7 & 83.13$\pm$0.27 & 72.80$\pm$0.31 & 80.99$\pm$0.42 & 69.75$\pm$0.63 & 88.15$\pm$0.74 & 80.74$\pm$0.71 \\
A6: LK only, $k=17$ & 1 MW & 19.187 & 80.8 & 83.83$\pm$0.37 & 73.75$\pm$0.43 & 80.96$\pm$0.90 & 69.79$\pm$1.24 & 88.56$\pm$0.52 & 81.36$\pm$0.72 \\
\textbf{A7/SAR (Ours), $k=13$} & \textbf{1 MW} & \textbf{19.160} & \textbf{74.9} & \textbf{83.89$\pm$0.25} & \textbf{73.88$\pm$0.23} & \textbf{81.09$\pm$0.62} & \textbf{69.89$\pm$0.79} & 88.52$\pm$0.61 & 81.26$\pm$0.75 \\
A8/SAR last-block only, $k=17$ & 1 MW & 19.062 & 76.5 & 82.83$\pm$0.25 & 72.57$\pm$0.49 & 80.58$\pm$0.42 & 69.22$\pm$0.49 & \textbf{88.91$\pm$0.31} & \textbf{81.54$\pm$0.37} \\
\bottomrule
\end{tabular}}
\end{table}

\begin{table}[H]
\centering
\scriptsize
\caption{Placement diagnostic. P0=A1 and P3=A7, with accuracy values identical to Table~\ref{tab:ablation}. P1/P2/P3 all use LK-Gate+SE with $k=13$ and differ only in insertion location. Params and GFLOPs are profiled on BUSI at $256\times256$ with one forward pass; P0/P3 complexity values are reused from the corresponding ablation profiles.}
\label{tab:placement}
\setlength{\tabcolsep}{2.6pt}
\resizebox{\linewidth}{!}{%
\begin{tabular}{l cc cc cc cc}
\toprule
\textbf{Placement} &
\textbf{Params} &
\textbf{GFLOPs} &
\multicolumn{2}{c}{\textbf{BUSI}} &
\multicolumn{2}{c}{\textbf{Nerve}} &
\multicolumn{2}{c}{\textbf{Kvasir}} \\
\cmidrule(lr){4-5}\cmidrule(lr){6-7}\cmidrule(lr){8-9}
& & &
\makecell{\textbf{Dice}\\\textbf{-AUC}} &
\makecell{\textbf{IoU}\\\textbf{-AUC}} &
\makecell{\textbf{Dice}\\\textbf{-AUC}} &
\makecell{\textbf{IoU}\\\textbf{-AUC}} &
\makecell{\textbf{Dice}\\\textbf{-AUC}} &
\makecell{\textbf{IoU}\\\textbf{-AUC}} \\
\midrule
P0: no SAR & 18.903 & 409.293 & 83.38$\pm$0.14 & 73.24$\pm$0.30 & 80.24$\pm$0.31 & 68.92$\pm$0.46 & 87.79$\pm$0.64 & 79.95$\pm$0.93 \\
P1: input-side & 18.904 & 409.335 & 83.23$\pm$0.38 & 73.06$\pm$0.42 & 80.44$\pm$0.50 & 69.06$\pm$0.74 & 87.51$\pm$0.63 & 79.56$\pm$0.91 \\
P2: pre-fusion & 19.143 & 409.293 & 83.34$\pm$0.12 & 73.16$\pm$0.18 & 80.32$\pm$0.21 & 69.00$\pm$0.29 & 88.50$\pm$0.18 & 80.94$\pm$0.26 \\
\textbf{P3: post-fusion} & \textbf{19.160} & \textbf{445.783} & \textbf{83.89$\pm$0.25} & \textbf{73.88$\pm$0.23} & \textbf{81.09$\pm$0.62} & \textbf{69.89$\pm$0.79} & \textbf{88.52$\pm$0.61} & \textbf{81.26$\pm$0.75} \\
\bottomrule
\end{tabular}}
\end{table}

\begin{table}[H]
\centering
\tiny
\caption{Paired bootstrap 95\% confidence intervals over matched seeds. Point estimates follow Tables~\ref{tab:ablation} and~\ref{tab:placement}. The study uses only three paired seeds, so narrow intervals indicate low observed seed-level variation. A2 is the raw three-support baseline.}
\label{tab:paired_ci}
\setlength{\tabcolsep}{2.0pt}
\resizebox{\linewidth}{!}{%
\begin{tabular}{l cc cc cc}
\toprule
\textbf{Comparison} & \makecell{\textbf{BUSI}\\$\Delta$ \textbf{Dice-AUC}} & \makecell{\textbf{BUSI}\\$\Delta$ \textbf{IoU-AUC}} & \makecell{\textbf{Nerve}\\$\Delta$ \textbf{Dice-AUC}} & \makecell{\textbf{Nerve}\\$\Delta$ \textbf{IoU-AUC}} & \makecell{\textbf{Kvasir}\\$\Delta$ \textbf{Dice-AUC}} & \makecell{\textbf{Kvasir}\\$\Delta$ \textbf{IoU-AUC}} \\
\midrule
A7/SAR vs A2: MultiverSeg raw 3-shot & \textbf{$+0.89$ {\footnotesize$[0.13,1.31]$}} & \textbf{$+1.11$ {\footnotesize$[0.31,1.51]$}} & $+0.97$ {\footnotesize$[-0.29,1.69]$} & $+0.93$ {\footnotesize$[-0.68,2.15]$} & \textbf{$+0.50$ {\footnotesize$[0.35,0.70]$}} & \textbf{$+0.94$ {\footnotesize$[0.75,1.14]$}} \\
A7/SAR vs A1: no recal. & \textbf{$+0.51$ {\footnotesize$[0.21,0.66]$}} & \textbf{$+0.64$ {\footnotesize$[0.30,0.98]$}} & $+0.85$ {\footnotesize$[-0.13,1.72]$} & $+0.97$ {\footnotesize$[-0.30,2.06]$} & $+0.73$ {\footnotesize$[-0.16,2.13]$} & $+1.31$ {\footnotesize$[-0.03,3.16]$} \\
A7/SAR vs A3/SAR, $k=17$ & $+0.04$ {\footnotesize$[-0.44,0.31]$} & $+0.09$ {\footnotesize$[-0.49,0.39]$} & $+0.03$ {\footnotesize$[-0.27,0.27]$} & $-0.05$ {\footnotesize$[-0.31,0.18]$} & $-0.11$ {\footnotesize$[-0.52,0.45]$} & $-0.12$ {\footnotesize$[-0.84,0.66]$} \\
A7/SAR vs A6: LK only, $k=17$ & $+0.06$ {\footnotesize$[-0.12,0.23]$} & $+0.13$ {\footnotesize$[-0.15,0.37]$} & $+0.13$ {\footnotesize$[-0.30,0.52]$} & $+0.10$ {\footnotesize$[-0.44,0.39]$} & $-0.04$ {\footnotesize$[-0.73,0.69]$} & $-0.10$ {\footnotesize$[-0.83,0.64]$} \\
A7/SAR vs A8/SAR last-block only & $+1.06$ {\footnotesize$[1.04,1.08]$} & $+1.31$ {\footnotesize$[1.02,1.53]$} & \textbf{$+0.51$ {\footnotesize$[0.35,0.82]$}} & \textbf{$+0.67$ {\footnotesize$[0.30,1.22]$}} & $-0.39$ {\footnotesize$[-0.82,0.24]$} & $-0.28$ {\footnotesize$[-0.62,0.40]$} \\
A7/SAR vs P1: input-side & \textbf{$+0.66$ {\footnotesize$[0.23,1.00]$}} & \textbf{$+0.82$ {\footnotesize$[0.39,1.08]$}} & $+0.65$ {\footnotesize$[-0.10,1.26]$} & \textbf{$+0.83$ {\footnotesize$[0.17,1.44]$}} & \textbf{$+1.01$ {\footnotesize$[0.89,1.09]$}} & \textbf{$+1.70$ {\footnotesize$[1.56,1.87]$}} \\
A7/SAR vs P2: pre-fusion & \textbf{$+0.55$ {\footnotesize$[0.26,0.94]$}} & \textbf{$+0.72$ {\footnotesize$[0.40,1.15]$}} & \textbf{$+0.77$ {\footnotesize$[0.36,1.23]$}} & \textbf{$+0.89$ {\footnotesize$[0.47,1.67]$}} & $+0.02$ {\footnotesize$[-0.56,0.53]$} & $+0.32$ {\footnotesize$[-0.49,0.88]$} \\
\bottomrule
\end{tabular}}
\end{table}

Table~\ref{tab:ablation} summarizes the controlled ablation rows. The main SAR row has higher mean Dice-AUC/IoU-AUC than the raw-support comparator on all three datasets, with paired differences reported in Table~\ref{tab:paired_ci}. The A7 vs A1 comparison isolates SAR under the same CST support representation. The single-component rows A4 (SE only) and A5 (LK only, $k=13$) are lower than A7 on BUSI and NerveUS-BP-present, suggesting that the combination of spatial gating and channel recalibration is useful; the three-seed study does not make this a definitive component-level significance claim.

Against A2, A7 gives $+0.89$ Dice-AUC $[0.13,1.31]$ and $+1.11$ IoU-AUC $[0.31,1.51]$ on BUSI. On NerveUS-BP-present, A7 has the highest Dice-AUC, but the interval includes zero: $+0.97$ Dice-AUC $[-0.29,1.69]$ and $+0.93$ IoU-AUC $[-0.68,2.15]$. On Kvasir-SEG, the corresponding differences are $+0.50$ Dice-AUC $[0.35,0.70]$ and $+0.94$ IoU-AUC $[0.75,1.14]$. A8 has the highest Kvasir mean and lower Kvasir standard deviation, but it is weaker on BUSI and NerveUS-BP-present. A7 is used as the main setting because it gives the best BUSI performance, the highest Nerve Dice-AUC, and a positive mean gain over A2 on Kvasir-SEG.

The narrow BUSI A7 vs A8 interval comes from the observed matched seed differences under only three paired seeds. The logged training times are $74.9$ and $89.5$ min/run for the main SAR and raw-support settings, respectively. The main SAR setting has more profiled single-pass GFLOPs ($445.783$ vs. $409.293$), so the lower logged training time in this implementation is not a FLOP-level speedup. The wall-clock difference comes from implementation overhead: SAR precomputes one CST once from the fixed 100-image training bank and reuses the resulting compact tensor, whereas the raw-support setting repeatedly loads and processes three supports with gradient-tracked computation, tensor movement, and backward-pass overhead.

\medskip\noindent\textbf{Input--pre-fusion--post-fusion placement diagnostic.}
Table~\ref{tab:placement} uses the same LK-Gate+SE configuration ($k=13$) for P1/P2/P3 and changes only insertion location. P3/A7 improves over P2 on BUSI and NerveUS-BP-present; on Kvasir it is comparable to P2 and below the A8 last-block-only diagnostic. These results favor late post-fusion recalibration, while the best exact insertion depth differs across datasets.

\section{Discussion and Conclusion}

Support-Aware Recalibration (SAR) is a compact late post-fusion recalibration block for one-click support-conditioned segmentation. The controlled ablations support SAR as a useful late recalibration step across BUSI, NerveUS-BP-present, and Kvasir-SEG, with the strongest evidence on BUSI and Kvasir-SEG and weaker three-seed evidence on NerveUS-BP-present. A7 is selected as the main SAR setting because it is the most consistent choice across datasets, while A3 and A8 remain kernel-size and insertion-depth diagnostics rather than separate main methods.

The CST is query-independent and built only from a fixed 100-image training support bank. It reduces repeated raw-support processing while providing a fixed foreground reference; SAR then recalibrates the late fused representation instead of relying on stronger prompts, test-time support selection, or test-set threshold tuning.

\medskip\noindent\textbf{Limitations.}
The confidence intervals use only three seeds under one frozen split, so they measure training-run variability, not split robustness. Table~\ref{tab:main} uses source-reported results, not a controlled external rerun, because SAM-family methods differ in resolution, prompt encoders, preprocessing, and training recipes. The study also covers only three foreground-present datasets at $256\times256$ resolution. More seeds, repeated splits, a paired permutation test, and a controlled MedSAM/SAMUS rerun would strengthen future versions.

\medskip\noindent\textbf{Code and data availability.}
Code, processed split lists, click lists, and trained weights will be released upon acceptance.

Overall, the controlled ablations favor late post-fusion recalibration for leakage-controlled one-click segmentation.

\end{document}